\documentclass[11pt]{article}
\usepackage{coling2020}
\usepackage{times}
\usepackage{url}
\usepackage{latexsym}
\usepackage{microtype}
\usepackage{amsmath}
\usepackage{amsfonts}
\usepackage{graphicx}
\usepackage{booktabs}
\usepackage{algorithm}
\usepackage{algorithmic}
\usepackage{multirow}
\usepackage{caption}
\usepackage{subcaption}
\usepackage{tabu}
\usepackage{makecell}
\usepackage{footnote}
\usepackage{footmisc}
\usepackage{color}
\usepackage{scalerel}
\usepackage{xcolor}
\usepackage{ulem}
\usepackage{varwidth}

\colingfinalcopy 

\title{Improving Variational Autoencoder for Text Modelling with Timestep-Wise Regularisation}

\author{Ruizhe Li\textsuperscript{$\spadesuit$},
Xiao Li\textsuperscript{$\spadesuit$}, 
Guanyi Chen\textsuperscript{$\heartsuit$},  
Chenghua Lin\textsuperscript{$\spadesuit$}\thanks{~~Corresponding author}~\\
\textsuperscript{$\spadesuit$}Department of Computer Science, University of Sheffield\\
\textsuperscript{$\heartsuit$}Department of Information and Computing Sciences, Utrecht University\\
\texttt{\{r.li, xiao.li, c.lin\}@sheffield.ac.uk, 
g.chen@uu.nl}}

\date{}

\begin{document}
\maketitle
\begin{abstract}
The Variational Autoencoder (VAE) is a popular and powerful model applied to text modelling to generate diverse sentences. However, an issue known as posterior collapse (or KL loss vanishing) happens when the VAE is used in text modelling, where the approximate posterior collapses to the prior, and the model will totally ignore the latent variables and be degraded to a plain language model during text generation. Such an issue is particularly prevalent when  RNN-based VAE models are employed for text modelling. In this paper, we propose a simple, generic architecture called Timestep-Wise Regularisation VAE (TWR-VAE), which can effectively avoid posterior collapse and can be applied to any RNN-based VAE models. The effectiveness and 
versatility of our model are demonstrated in different tasks, including language modelling and dialogue response generation.
\end{abstract}

\section{Introduction}\label{sec:introduction}
\blfootnote{
\hspace{-0.65cm}  
    This work is licensed under a Creative Commons 
    Attribution 4.0 International Licence.
    Licence details:
    \url{http://creativecommons.org/licenses/by/4.0/}.
    }
Variational Autoencoders (VAE)~\cite{kingma2013auto,rezende2014stochastic}, together with other deep generative models, including Generative Adversarial Networks~\cite{goodfellow2014generative} and autoregressive models~\cite{oord2018representation}, have attracted a mass of attention in the research community as they have shown their ability to learn compact representations from complex, high-dimensional unlabelled data. VAEs have been widely used in many NLP tasks, such as text modelling~\cite{bowman2016generating,yang2017improved,xu2018spherical,fang2019implicit,li2019stable}, style transfer~\cite{fang2019implicit}, and response generation~\cite{zhao2017learning,fang2019implicit}. In addition, VAEs are also useful to several downstream tasks, e.g., classification~\cite{xu2017variational,zhao2017learning,li2019dual,gururangan2019variational}, transfer learning~\cite{higgins2017darla}, etc.

However, there is a challenging optimisation issue of VAEs known as posterior collapse (a.k.a. KL loss vanishing), where the variational posterior collapses to the prior and the latent variable is ignored by the model during generation~\cite{bowman2016generating}. This is particularly prevalent when employing VAE-RNN architectures for text modelling. 
When posterior collapse happens, the decoder will be downgraded to a simpler language model and the VAE cannot learn good latent representations of data~\cite{sonderby2016train,yang2017improved}. Different strategies have been proposed to address this issue, such as annealing the KL term in the VAE loss function~\cite{bowman2016generating,sonderby2016train,fu2019cyclical}, replacing the recurrent decoder with convolutional neural networks (CNNs)~\cite{yang2017improved,semeniuta2017hybrid}, using a sophisticated prior distribution such as the von Mises-Fisher (vMF) distribution~\cite{xu2018spherical}; and adding mutual information into the VAE objectives~\cite{phuong2018mutual}. While the aforementioned strategies have shown effectiveness in tackling the posterior collapse issue to some extent, they either require careful engineering between the reconstruction loss and the KL loss~\cite{bowman2016generating,sonderby2016train,fu2019cyclical}, or designing more sophisticated model structures~\cite{yang2017improved,semeniuta2017hybrid,xu2018spherical,phuong2018mutual}.

In this paper, we propose a simple and robust architecture called Timestep-Wise Regularisation VAE (TWR-VAE), which can effectively alleviate the VAE posterior collapse issue  in text modelling. Existing VAE-RNN models for text modelling only impose KL regularisation on the latent variable of the RNN encoder at the final timestep, forcing the latent variable to be close to a Gaussian prior.  In contrast, our TWR-VAE imposes KL regularisation on 
the latent variables of every timestep of the RNN encoder, which we dub \textit{timestep-wise regularisation}.
We hypothesise that timestep-wise regularisation is crucial to avoid  posterior collapse and to learn good representations of data, and allows a more robust  model learning process. 
In addition, the proposed timestep-wise regularisation strategy is generic and in theory can be applied to any existing VAE-RNN models, e.g., vanilla RNN and GRU-based VAE models. 
TWR-VAE shares some similarity with existing VAE-RNN models, where the input to the decoder is the latent variable sample from the variational posterior at the final timestep of the encoder.   
While this is a reasonable design choice, we also explore two model variants of TWR-VAE, namely, TWR-VAE$_\text{mean}$ and TWR-VAE$_\text{sum}$. At each time step, both model variants sample a latent variable from the timestep dependent variational posterior of the encoder. TWR-VAE$_\text{mean}$ averages the sampled latent variables which is then used as input to the decoder, whereas TWR-VAE$_\text{sum}$ performs vector addition on the sampled latent variables instead. 

To demonstrate the effectiveness of our method, we select a number of strong  baseline models and conduct comprehensive evaluations in two benchmark tasks involving five public datasets.  
For the language modelling task, experimental results show that our TWR-VAE model can effectively 
alleviate the posterior collapse issue and consistently give better predictive performance than all the baselines as evidenced by both quantitative (e.g., negative log likelihood and perplexity) and qualitative evaluation. 
For the dialogue response generation task, our model yields better or comparable performance to the state-of-the-art baselines based on three evaluation metrics (i.e. BLEU~\cite{zhao2017learning}, BOW embedding~\cite{liu2016not} and Dist~\cite{liu2016not}). Manual examination also shows that the dialogue responses generated by our model are more diverse and contentful than the baselines, as well as being simpler in model design. 
Our two model variants also show comparable performance to the best baseline, although not as strong as TWR-VAE.

In summary, the contribution of our paper are three-fold: (1) we propose a simple and robust method, which can effectively alleviate the posterior collapse issue of VAE via timestep-wise regularisation; 
(2) our approach is generic  which can be applied to any RNN-based VAE models; (3) our approach outperforms the state-of-art on language modelling and yields better or comparable performance on dialogue response generation. The code of TWR-VAE is available at: \url{https://github.com/ruizheliUOA/TWR-VAE}.

\section{Related Work}

Variational autoencoder~\cite{kingma2013auto,rezende2014stochastic} yields great performance when it was applied to image generation~\cite{razavi2019generating}, facial attribute style transfer~\cite{hou2017deep,klys2018learning,icml2020_1832}, etc.
It also has been applied to many natural language processing tasks, including text generation~\cite{bowman2016generating,fang2019implicit,zhu-etal-2020-batch}, dialogue response generation~\cite{serban2017hierarchical,zhao2017learning,park2018hierarchical,gu2019dialogwae,fang2019implicit}, and style transfer~\cite{john2019disentangled,fang2019implicit,xu2019variational}
For all these applications, there is a common issue called posterior collapse (or KL loss vanishing)~\cite{bowman2016generating}.

Several different types of methods were proposed to address this issue. KL annealing is the most common and basic solution used in almost all  works~\cite{bowman2016generating,sonderby2016train,semeniuta2017hybrid,he2019lagging,fu2019cyclical,fang2019implicit}.  Another type of approaches attempt to weaken the decoder of VAE to avoid posterior collapse,
such as introducing word dropout and historyless decoding into the decoder~\cite{bowman2016generating}, replacing the decoder with different CNNs~\cite{yang2017improved,semeniuta2017hybrid}, and adding skip connections in the decoder~\cite{dieng2019avoiding}. Others tried to solve this issue by introducing new regularisers~\cite{zhao2017infovae,goyal2017z,tolstikhin2018wasserstein}, using more sophisticated prior distributions~\cite{tomczak2018vae,xu2018spherical}, etc.

More recently, Fu et al.~\shortcite{fu2019cyclical} used a cyclical annealing schedule to alleviate the KL loss vanishing issue. He et al.~\shortcite{he2019lagging} proposed a lagging inference network to update the encoder multiple times before a single decoder update to address the issue from the perspective of training dynamics.
Zhu et al.~\shortcite{zhu-etal-2020-batch} applied the batch normalisation to the parameters of the approximate posterior and ensured that the lower bound of the expectation of the KL is positive to avoid posterior collapse.
In contrast, our approach only imposes the KL regularisation on timestep-wise latent variables in the encoder, which is simpler without changing the VAE training mode, introducing more complicated posterior distributions or adding a KL annealing as a warm-up setup.

\begin{figure*}
\centering
    \includegraphics[width=\textwidth]{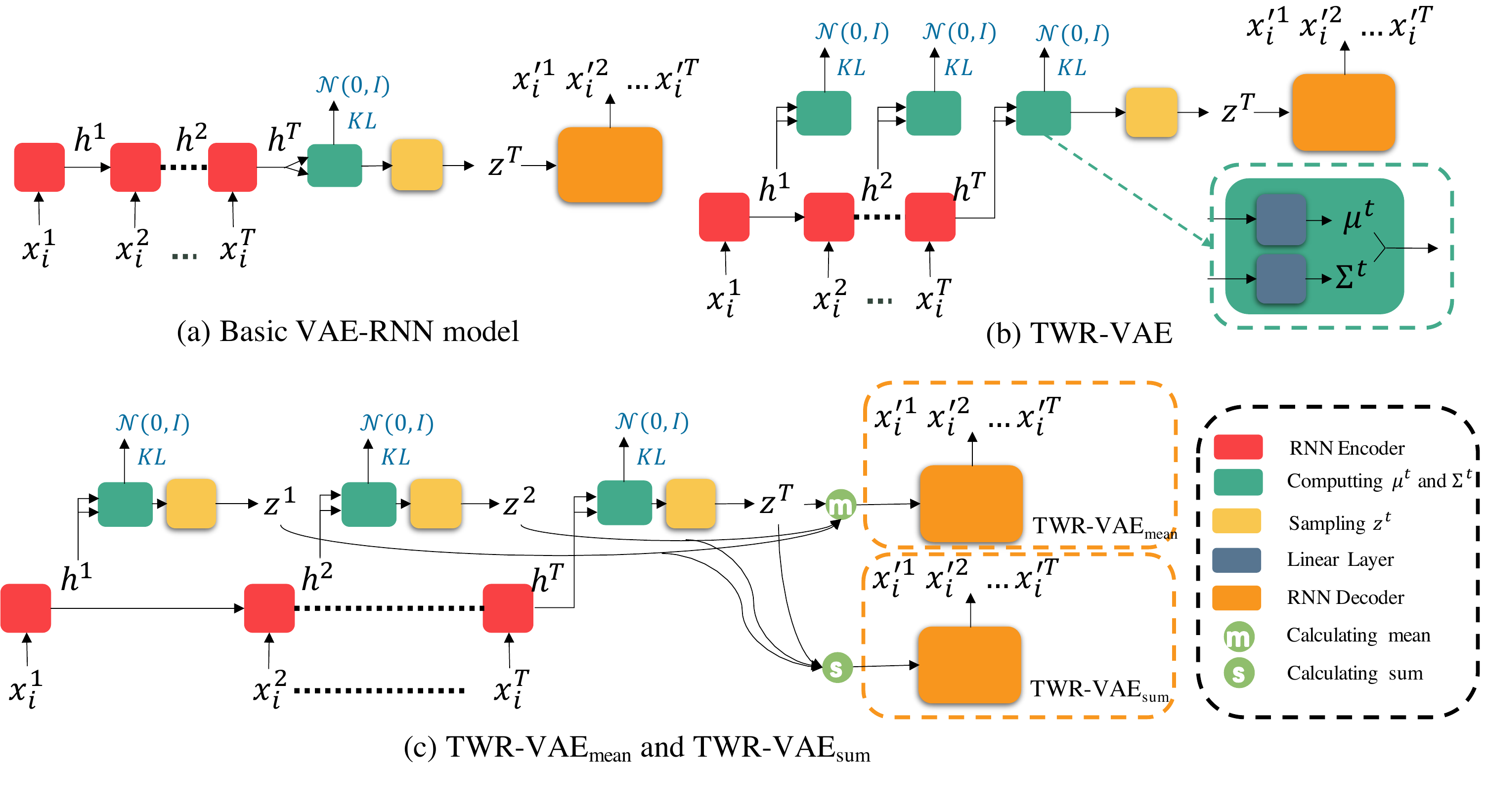}
    \caption{Architectures of the proposed TWR-VAE models and the basic VAE-RNN model.}
    \label{fig:model}
\end{figure*}

\section{Methodology}

In this section, we introduce the proposed Timestep-Wise Regularisation VAE (TWR-VAE) model as well as its two model variants. We briefly introduce the background of VAE before describing the technical details of the proposed models. 

\subsection{Background of VAE}

A variational autoencoder is a generative model, which is designed to generate data via a latent variable $\mathbf{z}$. For a dataset $\mathbf{X}=\{\mathbf{x}_i\}_{i=1}^{N}$ with $N$ i.i.d. data, there are two steps in the data generation process: (1) a latent variable $\mathbf{z}$ is sampled from a prior distribution $P_{\boldsymbol{\theta}}(\mathbf{z})$; (2) a data $\mathbf{x}_i$ is generated from the conditional distribution $P_{\boldsymbol{\theta}}(\mathbf{x}_i|\mathbf{z})$. We need to optimise the marginal likelihood $P_{\boldsymbol{\theta}}(\mathbf{x}_i)=\int{P_{\boldsymbol{\theta}}(\mathbf{z})P_{\boldsymbol{\theta}}(\mathbf{x}_i|\mathbf{z})d\mathbf{z}}$ using VAE. However, both of the marginal likelihood $P_{\boldsymbol{\theta}}(\mathbf{x}_i)$ and the true posterior distribution $P_{\boldsymbol{\theta}}(\mathbf{z} | \mathbf{x}_i)=P_{\boldsymbol{\theta}}(\mathbf{x}_i | \mathbf{z}) P_{\boldsymbol{\theta}}(\mathbf{z}) / P_{\boldsymbol{\theta}}(\mathbf{x}_i)$ are intractable. In order to train VAE, an encoder $Q_{\boldsymbol{\phi}}(\mathbf{z}|\mathbf{x}_i)$ is used to approximate the true posterior $P_{\boldsymbol{\theta}}(\mathbf{z} | \mathbf{x}_i)$. In this way, a data $\mathbf{x}_i$ is encoded as a distribution of $\mathbf{z}$ via the encoder $Q_{\boldsymbol{\phi}}(\mathbf{z}|\mathbf{x}_i)$ and the latent code $\mathbf{z}$ is fed into the decoder $P_{\boldsymbol{\theta}}(\mathbf{x}_i|\mathbf{z})$ to decode a distribution over some values of $\mathbf{x}_i$.

In general, the VAE is trained to maximise the marginal log likelihood $\log P_{\boldsymbol{\theta}}\left(\mathbf{x}_{1},\ldots, \mathbf{x}_{N}\right)=\sum_{i=1}^{N} \log P_{\boldsymbol{\theta}}\left(\mathbf{x}_{i}\right)$  for the whole training dataset. 
This is essentially equivalent to maximising the following  evidence lower bound (ELBO)\footnote{See Appendix~\ref{AP:A} for the full  derivation.}, which consists of two terms~\cite{kingma2013auto}: \begin{align}\label{eq:ELBO}
    \mathcal{L}(\boldsymbol{\theta},\boldsymbol{\phi};\mathbf{x}_i)&=\mathbb{E}_{Q_{\boldsymbol{\phi}}(\mathbf{z}|\mathbf{x}_i)}[\log P_{\boldsymbol{\theta}}(\mathbf{x}_i|\mathbf{z})]  -D_\text{KL}\left(Q_{\boldsymbol{\phi}}(\mathbf{z}|\mathbf{x}_i)\|P(\mathbf{z})\right)\,.
\end{align}
The first term is the expected reconstruction error indicating how well the model can reconstruct data given a latent variable.  The the second term is the KL-divergence of the approximate posterior from prior, i.e., a regularisation pushing the learned posterior to be as close to the prior as possible.
The basic VAE-RNN model (Figure~\ref{fig:model}(a)) follows the aforementioned ELBO (i.e. Eq.~\ref{eq:ELBO}). As the architecture of the encoder is a RNN, a latent variable (denoted as $\mathbf{z}^{\scriptscriptstyle T}$) is sampled from the variational posterior at the final timestep $T$, and then  $\mathbf{z}^{\scriptscriptstyle T}$ is used as the input to the decoder. Therefore, the ELBO of a basic VAE-RNN model becomes:
\begin{align}\label{eq:RNN-ELBO}
    \mathcal{L}(\boldsymbol{\theta},\boldsymbol{\phi};\mathbf{x}_i)_{\text{basic}}&=\mathbb{E}_{Q_{\boldsymbol{\phi}}(\mathbf{z}^{\scaleto{T}{3.5pt}}|\mathbf{x}_i)}[\log P_{\boldsymbol{\theta}}(\mathbf{x}_i|\mathbf{z}^{\scriptscriptstyle T})]  -D_\text{KL}\left(Q_{\boldsymbol{\phi}}(\mathbf{z}^{\scriptscriptstyle T}|\mathbf{x}_i)\|P(\mathbf{z}^{\scriptscriptstyle T})\right)\,.
\end{align}
Note that the total number of timestep $T$ is also the length of the input sentence.  
As discussed, optimising ELBO (in Eq.~\ref{eq:RNN-ELBO}) is prone to posterior collapsing to the prior~\cite{bowman2016generating}. This phenomenon happens when the second term of Eq.~\ref{eq:RNN-ELBO} would approach to its global minimum when $Q_{\boldsymbol{\phi}}(\mathbf{z}^{\scriptscriptstyle T}|\mathbf{x}_i) = P(\mathbf{z}^{\scriptscriptstyle T})$, which results that $\mathbf{x}$ and $\mathbf{z}^{\scriptscriptstyle T}$ are two independent variables. As a result, the decoder (i.e., the reconstruction term) no longer depends on $\mathbf{z}^{\scriptscriptstyle T}$ and it fits the training data as a plain language model.

\subsection{Variational Autoendoder with Timestep-Wise Regularisation (TWR-VAE)} \label{sec:hr-vae}

In this section, we introduce the proposed Timestep-Wise Regularisation (TWR-VAE) model, a general architecture which can effectively mitigate the posterior collapse issue frequently observed in the VAE models with RNN-based backbone. 

Our model design is motivated by one noticeable defect shared by the VAE-RNN based models in previous works~\cite{bowman2016generating,yang2017improved,xu2018spherical,dieng2019avoiding}. That is, the general architecture of all these models, as shown in Figure~\ref{fig:model}(a), only impose a standard normal distribution prior on the last hidden state of the RNN encoder, which potentially leads to learning a suboptimal representation of the latent variable. 
In addition, to avoid  posterior collapsing, it is important to learn good latent representations of data at the early stage of  decoder training, so that the decoder can easily adopt them to generate controllable observations~\cite{fu2019cyclical}.  
Our hypothesis is that to learn a good representation of data, it is crucial to impose the standard normal prior on the hidden states of all timesteps of the RNN-based encoder, which  will allow a better regularisation of the model learning process especially during the early stages. 

The architecture of the proposed TWR-VAE model is shown in Figure~\ref{fig:model}(b), which is implemented using a one-layer LSTM for both the encoder and decoder. 
For each timestep $t$, 
we feed the hidden state $\mathbf{h}^t$ into two linear transformation layers for estimating  $\boldsymbol{\mu}^t$ and  $\boldsymbol{\Sigma}^t$, which are parameters of the variational posterior, i.e., a normal distribution corresponding to the $\mathbf{h}^t$.
We then impose KL regularisation on all timestep-wise variational posteriors rather than  posterior of the last timestep.
Formally, given input  $\mathbf{X}=\{\mathbf{x}_i\}_{i=1}^{N}$, the ELBO of our model for each data pint $\mathbf{x}_i$ is defined as:
\begin{align}\label{eq:our_ELBO}
    \mathcal{L}(\boldsymbol{\theta},\boldsymbol{\phi};\mathbf{x}_i)_{\text{TWR}} &= \mathbb{E}_{Q_{\boldsymbol{\phi}}(\mathbf{z}^{\scaleto{T}{3.5pt}}|\mathbf{x}_i)}[\log P_{\boldsymbol{\theta}}(\mathbf{x}_i|\mathbf{z}^{\scriptscriptstyle T})]  - \frac{1}{T} \sum_{t=1}^{T}D_\text{KL}(Q_{\boldsymbol{\phi}}(\mathbf{z}^t | \mathbf{x}_i^{1:t}) \| P(\mathbf{z}^t))\,,
\end{align}
where $T$ is the length of the input sentence, $\boldsymbol{\theta}$ and $\boldsymbol{\phi}$ are the parameters for the decoder and the encoder, respectively. 
Note that TWR-VAE is similar to existing VAE-RNN models~\cite{xu2018spherical,fu2019cyclical,he2019lagging}, which passes a single $\mathbf{z}^{\scriptscriptstyle T}$ at the final timestep to the decoder. However, there is a crucial difference that while existing models only impose KL regularisation on the last timestep, TWR-VAE imposes timestep-Wise KL regularisation and \textit{average the KL loss over all timesteps}, i.e., the second term of Eq.~\ref{eq:our_ELBO}. 
Such a strategy allows more robust model learning and can effectively mitigate posterior collapse (see \S\ref{sec:experiment} Experiment for detailed discussion). Compared to the HR-VAE of Li et al.,~\shortcite{li2019stable}, our model does not concatenate the cell state of the encoder at each timestep and the dimension of the latent variable of TWR-VAE is only 32, whereas for HR-VAE the dimension is 512 which is much larger. This enables the proposed TWR-VAE model to have fewer parameters than the HR-VAE. In addition, the training speed of the TWR-VAE is six times faster than the HR-VAE by paralleling the timestep-wise KL regularisation.

Following~\newcite{kingma2013auto}, a reparameterisation trick is used to enable the timestep-wise latent variable sampling differentiable. During the gradients optimisation of $\boldsymbol{\theta}$ and $\boldsymbol{\phi}$, we use Monte Carlo method~\cite{metropolis1949monte} to construct a Monte Carlo estimator, which can obtain unbiased gradients of $\boldsymbol{\theta}$ and $\boldsymbol{\phi}$ (see Appendices~\ref{AP:B} and~\ref{AP:C} for the detailed derivation):
\begin{align}\label{eq:gradient_theta_phi_HR-VAE-ELBO}
    \nabla_{\boldsymbol{\theta},\boldsymbol{\phi}}\mathcal{L}(\boldsymbol{\theta},\boldsymbol{\phi};\mathbf{x}_i)_\text{TWR}&\simeq \frac{1}{M}\sum_{m=1}^{M}\nabla_{\boldsymbol{\theta},\boldsymbol{\phi}}\left(\log P_{\boldsymbol{\theta}}(\mathbf{x}_i|\mathbf{z}_{m}^{\scriptscriptstyle T})-\frac{1}{T} \sum_{t=1}^{T}\log\frac{Q_{\boldsymbol{\phi}}(\mathbf{z}_{m}^t|\mathbf{x}_i^{1:t})}{P(\mathbf{z}_{m}^t)}\right) \nonumber \\ & \quad \text{where} \quad \mathbf{z}_{m}^t=Q_{\boldsymbol{\phi}}(\mathbf{z}_{m}^t|\mathbf{x}_i^{1:t})\,,
\end{align}
Here $M$ indicates the total number of times that we randomly sample $\mathbf{z}^{t}_m$ ($m \in [1:M]$) from the $Q_{\boldsymbol{\phi}}(\mathbf{z}_{m}^t|\mathbf{x}_i^{1:t})$ for approximation.

\subsection{TWR-VAE$_\text{mean}$ and TWR-VAE$_\text{sum}$}

In TWR-VAE, the input to the decoder is the latent variable sample from the variational posterior at the final timestep of the encoder.   
While this is a reasonable design choice, we also explore two model variants of TWR-VAE, namely, TWR-VAE$_\text{mean}$ and TWR-VAE$_\text{sum}$ (see Figure~\ref{fig:model}(c)). At each time step, both model variants sample a latent variable from the timestep dependent variational posterior of the encoder. 

For TWR-VAE$_\text{mean}$, the timestep-wise latent variables $\{\mathbf{z}^t\}_{t=1}^T$ are sampled first and then they are averaged before feeding to the decoder. This leads to a different  reconstruction loss of TWR-VAE$_\text{mean}$ compared to TWR-VAE (Eq.~\ref{eq:our_ELBO}):
\begin{equation}
    \mathbb{E}[\log P_{\boldsymbol{\theta}}(\mathbf{x}_i|\frac{1}{T} \sum_{t=1}^{T}\mathbf{z}^{t})] \qquad \text{where} \: \mathbf{z}^{t}\sim Q_{\boldsymbol{\phi}}(\mathbf{z}^{t}|\mathbf{x}_i^{1:t})
\end{equation}

For TWR-VAE$_\text{sum}$, it performs vector addition on the sampled latent variables  $\{\mathbf{z}^t\}_{t=1}^T$ instead and the corresponding reconstruction loss is:
\begin{equation}
    \mathbb{E}[\log P_{\boldsymbol{\theta}}(\mathbf{x}_i|\sum_{t=1}^{T}\mathbf{z}^{t})] \qquad \text{where} \: \mathbf{z}^{t}\sim Q_{\boldsymbol{\phi}}(\mathbf{z}^{t}|\mathbf{x}_i^{1:t})
\end{equation}

For both TWR-VAE$_\text{mean}$ and TWR-VAE$_\text{sum}$, their KL loss term is the same as TWR-VAE, i.e., $- \frac{1}{T} \sum_{t=1}^{T}D_\text{KL}(Q_{\boldsymbol{\phi}}(\mathbf{z}^t | \mathbf{x}_i^{1:t}) \| P(\mathbf{z}^t))$.

\begin{table}[tb]
  \centering \small
  \begin{tabular}{lcccc}
    \Xhline{2.5\arrayrulewidth}
    Dataset & Train & Dev. & Test & Vocab. \\
    \hline
    PTB   & 42,068 & 3,370 & 3,761 & 9.95K \\
    Yelp15 & 100,000 & 10,000 & 10,000 &19.76K \\
    Yahoo & 100,000 & 10,000 & 10,000 & 19.73K\\
     SW & 2,316 & 60 & 62 & 20K\\
     DD & 11,118 & 1,000 & 1,000 & 22K\\
    \Xhline{2.5\arrayrulewidth}
     &  &  &  & \\
  \end{tabular}
  \caption{The statistics of the PTB, Yelp 2015, Yahoo, SW and DD datasets.}
  \label{T:datasets}
\end{table}

\section{Experiment}\label{sec:experiment}

\subsection{Language Modelling}\label{sec:lanuage modelling}

We evaluate our TWR-VAE model on three public benchmark datasets, namely, Penn Treebank (PTB)~\cite{marcus1993building}, Yelp15~\cite{yang2017improved}, and Yahoo~\cite{zhang2015character}, which have been widely used in previous work for text modelling~\cite{bowman2016generating,kim2018semi,fu2019cyclical,he2019lagging,zhu-etal-2020-batch}. The statistics of the datasets are summarised in Table~\ref{T:datasets}. We represent input data with 512-dimensional word2vec embeddings~\cite{mikolov2013distributed} and set the dimension of the hidden layers of both one-layer encoder and decoder to 256. Appendix~\ref{AP:D} shows more details.

We compare our TWR-VAE model with five strong baselines\footnote{\textbf{VAE-LSTM}: \url{https://github.com/timbmg/Sentence-VAE}; \textbf{SA-VAE}: \url{https://github.com/harvardnlp/sa-vae}; \textbf{Cyclical VAE}: \url{https://github.com/haofuml/cyclical_annealing}; \textbf{Lagging VAE}:  \url{https://github.com/jxhe/vae-lagging-encoder}; 
\textbf{BN-VAE}: \url{https://github.com/valdersoul/bn-vae}}:
\textbf{VAE-LSTM}: A VAE with LSTM and with KL annealing for tackling the posterior collapse issue~\cite{bowman2016generating}; (2) \textbf{SA-VAE}: A VAE using stochastic variational inference to refine the variational parameters initialised by Amortized variational inference~\cite{kim2018semi}; (3) \textbf{Cyclical VAE}: A VAE employing cyclical annealing to alleviate the posterior collapse issue~\cite{fu2019cyclical}; (4) \textbf{Lagging VAE}: A VAE updating the encoder more times than updating the decoder~\cite{he2019lagging}; (5) \textbf{BN-VAE}: A VAE utilising  Batch Normalisation for the KL distribution~\cite{zhu-etal-2020-batch}.

\begin{table*}[tb]
  \centering
  \small
  \begin{tabular}{l|cccc|cccc|cccc}
  \toprule
  \multirow{2}{*}{Model} &
  \multicolumn{4}{c|}{PTB} &
  \multicolumn{4}{c|}{Yelp15} &
  \multicolumn{4}{c}{Yahoo}\\ 
   
     & NLL$\downarrow$ & PPL$\downarrow$ & MI$\uparrow$ & KL & NLL$\downarrow$ & PPL$\downarrow$ & MI$\uparrow$ & KL & NLL$\downarrow$ & PPL$\downarrow$ & MI$\uparrow$ & KL \\
     \midrule
VAE-LSTM & 101.2 & 101.4 & 0.0 & 0.0 & 357.9& 40.6 & 0.0 & 0.0 & 328.6& 61.2& 0.0 & 0.0 \\
SA-VAE & 101.0 & 100.7 & 0.8 & 1.3 & 355.9 & 39.7 & 2.8 & 1.7 & 327.2& 60.2& 2.7 & 5.2 \\
Cyc-VAE & 102.8 & 109.0 & 1.3 & 1.4 & 359.5 & 41.3 & 1.0 & 2.0 & 330.6 & 65.3 & 2.0 & 2.1 \\
Lag-VAE & 100.9 & 99.8 & 0.8 & 0.9 & 355.9 & 39.7 & 2.4 & 3.8 & 326.7& 59.8& 2.9 & 5.7 \\
BN-VAE (0.7) & 100.2 & 96.9 & \textbf{5.5} & 7.2 & 355.9 & 39.7 & \textbf{7.4} & 9.1 & 327.4 & 60.2 & \textbf{7.4} & 8.8 \\
\midrule
TWR-VAE$_\text{sum}$ & 96.7 & 63.2 & 3.7 & 5.9 & 378.3 & 47.4 & 3.8 & 3.9 & 345.6 & 71.1 & 3.7 & 3.8\\
TWR-VAE$_\text{mean}$& 95.6 & 60.4 & 3.9 & 4.9 & 361.7 & 40.0 & 3.9& 3.5 & 324.8 & 55.0 & 4.1 & 4.8 \\
TWR-VAE & \textbf{86.6} & \textbf{40.9} & 4.1 & 5.0 & \textbf{344.3} & \textbf{33.5} &4.1& 3.1 & \textbf{317.3} & \textbf{50.2} & 4.1 & 3.3\\

\bottomrule
  \end{tabular}
  \caption{Language modelling results of all baselines and our models on the PTB, Yelp15 and Yahoo test datasets. The results of all baselines are reported based on~\cite{li2019surprisingly,zhu-etal-2020-batch}. $\downarrow$ denotes lower the better and $\uparrow$ higher the better.}
  \label{T:NLL_KL}
\end{table*}

We report the performance on four metrics:  negative log likelihood (NLL), perplexity (PPL), KL-divergence which measures the distance between two probability distributions,
and the mutual information of the input $\mathbf{x}$ and the latent variable $\mathbf{z}$, which measures how much information of $\mathbf{x}$ is obtained by $\mathbf{z}$.
Following~\newcite{dieng2019avoiding} and~\newcite{he2019lagging}, the mutual information is formulated as 
$I(\mathbf{x},\mathbf{z})=\mathbb{E}_{\mathbf{x}}[D_\text{KL}\left(Q_{\phi}(\mathbf{z}^{\scriptscriptstyle T}|\mathbf{x})\|P(\mathbf{z}^{\scriptscriptstyle T})\right)]-D_\text{KL}(Q_{\phi}(\mathbf{z}^{\scriptscriptstyle T})\|P(\mathbf{z}^{\scriptscriptstyle T}))$, where $Q_{\phi}(\mathbf{z}^{\scriptscriptstyle T})$ is an aggregated posterior and $D_\text{KL}(Q_{\phi}(\mathbf{z}^{\scriptscriptstyle T})\|P(\mathbf{z}^{\scriptscriptstyle T}))$ is the KL divergence between the aggregated posterior and the prior estimated by Monte Carlo estimators (see Appendix~\ref{AP:E} for the whole derivation).

\noindent\textbf{Results.}~~As depicted in Table~\ref{T:NLL_KL},
our TWR-VAE outperforms all baselines on all datasets. Compared to the strongest baseline BN-VAE, our model reduces NLL by 11.8 and PPL by 24.1 on average across three datasets, showing superior performance in reconstructing input sentences.
As shown in Table~\ref{T:NLL_KL}, the two variants of TWR-VAE also yields better performance to the baselines. For instance,  TWR-VAE$_\text{mean}$ outperforms all baselines on PTB and Yahoo datasets and yield comparable results to BN-VAE on Yelp.
This shows the effectiveness of our strategy of regularising timestep-wise  variational posteriors.

\noindent \textbf{Model generalisability and Ablation studies.}~~We also evaluate the model's generalisability by looking at how well our timestep-wise regularisor works in different RNN architectures.
To this end, we tested  Basic-VAE$_\text{RNN}$ and Basic-VAE$_\text{GRU}$ (i.e., vanilla RNN and GRU model with KL annealing ), as well as  TWR-VAE$_\text{RNN}$ and TWR-VAE$_\text{GRU}$ (vanilla RNN and GRU with the timestep-wise regularisor). 
Experimental results in Table~\ref{T:ablation} show that our TWR models outperform the corresponding basic models on all evaluation metrics, regardless the encoder architecture. This shows the generalisability of our proposed architecture.

In addition, to understand how the proportion of timesteps that are imposed with KL regularisation impacts the performance of our model,  we run a battery of experiments with varying  proportion settings. Concretely, we impose KL regularisation on the last 25\%, 50\%, and 75\% timesteps of the encoder of TWR-VAE, respectively.
(\textbf{NB}: the KL regularisation is imposed on the final timestep for all model variants).
The results in Table~\ref{T:ablation} show that  TWR-VAE$_{\text{LSTM-last25}}$ has the lowest performance on NLL and PPL and the performance goes up along with higher proportion of timesteps being  imposed with KL regularisation.
In addition, when comparing these three model variants with the baseline VAE-LSTM (which only imposes the KL regularisation on the final timestep), our models can effectively mitigate posterior collapse. 
This observation embodies that imposing the KL regularisation on earlier timesteps is an effective strategy for mitigating  posterior collapse. Moreover, the more timesteps we impose the KL regularisation on, the better performance the model can yield (in terms of NLL and PPL).

\begin{table}[t]
    \centering \small
    \begin{tabular}{l|cccc|cccc}
  \toprule
  \multicolumn{1}{c|}{\multirow{2}{*}{Model}} &
  \multicolumn{4}{c|}{Yelp15} &
  \multicolumn{4}{c}{Yahoo}\\ 
   
     & NLL$\downarrow$ & PPL$\downarrow$ & MI$\uparrow$ & KL & NLL$\downarrow$ & PPL$\downarrow$ & MI$\uparrow$ & KL \\
     \midrule
Basic-VAE$_{\text{RNN}}$ & 399.2 & 58.7 & 0.0 & 0.0 & 363.9 & 89.1 & 0.0 & 0.1\\
TWR-VAE$_{\text{RNN}}$ &  395.4& 56.4 & 3.9 & 0.5 &363.0 & 88.2 & 4.1 & 0.6 \\
Basic-VAE$_{\text{GRU}}$ & 389.6 & 53.2 & 0.6 & 0.6 & 355.0 & 79.9 & 2.3 & 2.6\\ 
TWR-VAE$_{\text{GRU}}$ &  360.9 & 39.7 & \textbf{4.2} & 3.3 & 336.9 & 63.9 & \textbf{4.2} & 3.7 \\\midrule
TWR-VAE$_{\text{LSTM-last25}}$ &  360.4 & 39.5 & 4.1 & 8.3 & 338.2 & 64.9 & \textbf{4.2} & 8.4 \\
TWR-VAE$_{\text{LSTM-last50}}$ &  356.2 & 37.9 & 4.1 & 5.1 & 331.7 & 59.9 & \textbf{4.2} & 5.3 \\
TWR-VAE$_{\text{LSTM-last75}}$ &  352.6 & 36.5 & 4.1 & 3.7 & 321.0 & 52.5 & 4.1 & 4.1 \\
TWR-VAE &  \textbf{344.3} & \textbf{33.5} & 4.1& 3.1 & \textbf{317.3} & \textbf{50.2} & 4.1 & 3.3 \\

\bottomrule
  \end{tabular}
  \caption{Ablation study results of all variants of our model on the Yelp15 and Yahoo test datasets.}
  \label{T:ablation}
\end{table}

\begin{table*}[tb] 
\small
  \centering 
  \begin{tabular}{c|l|p{12.5cm}} 
    \Xhline{2.5\arrayrulewidth}
    \multirow{2}{*}{Yelp15} & Input 1 & this is the worst restaurant experience i 've ever had ! not only is this place super slow in service but the food was not fresh !\\
    &Input 2 & i went to this place last month with my best friend and the food was good i love the coffee designs and the service was friendly .\\ \hline
    \multirow{6}{*}{\rotatebox[origin=c]{90}{BN-VAE}} &
    $\alpha=0$ & this place the worst restaurant i i have ever had . i only was the restaurant a overpriced , the , the food is not good and i\\
    &$\alpha=0.2$&this place joke ! the food was ok the was horrible . i ask for drink and came back to me . i will go back .\\
    &$\alpha=0.4$&this place joke ! the food was good horrible . i ask for a drink and check on me . i ask for a drink and check on me . \\
    &$\alpha=0.6$&i was try this place. disappointed . the food was not good it was just ok . the service was good the food was not price .\\
    &$\alpha=0.8$&i went lunch and the chicken and waffles . the food was good the service was horrible . i will go back .\\
    &$\alpha=1$&i went here this place for night and my family friend and i food was great . had the atmosphere and and the service was great . i\\
    \Xhline{2.5\arrayrulewidth}
    \multirow{6}{*}{\rotatebox[origin=c]{90}{TWR-VAE}} &$\alpha=0$&this is the worst restaurant i 've ever been ! service only was we restaurant was slow service but the food was not fresh !\\
    &$\alpha=0.2$&i love this place the food was very slow ! service is always slow and the food is not a good value so this was not my first choice .\\
    &$\alpha=0.4$&i have never been in this restaurant before the food was just ok and the service is very slow ! i will not continue to go back to this place .\\
    &$\alpha=0.6$&i have been here a few times now and the food was good ! ! ! the food is good and i would recommend to and return \\
    &$\alpha=0.8$&i went here this past weekend to see how good the food was and my husband had the same thing i would recommend for the price . \\
    &$\alpha=1$&i went to this place for night and my family friend and the food was good and would the service and the service was friendly . \\
    \Xhline{2.5\arrayrulewidth}
  \end{tabular}
  \caption{An example of interpolating the latent representation of two input sentences using BN-VAE and TWR-VAE in Yelp15 testset (see the example of Yahoo testset in appendix~\ref{AP:G}).}
  \label{T:sentence_yelp}
\end{table*}

\begin{figure*}[tb]
    \centering
    \begin{subfigure}[t]{0.5\textwidth}
        \centering
        \includegraphics[scale=0.5]{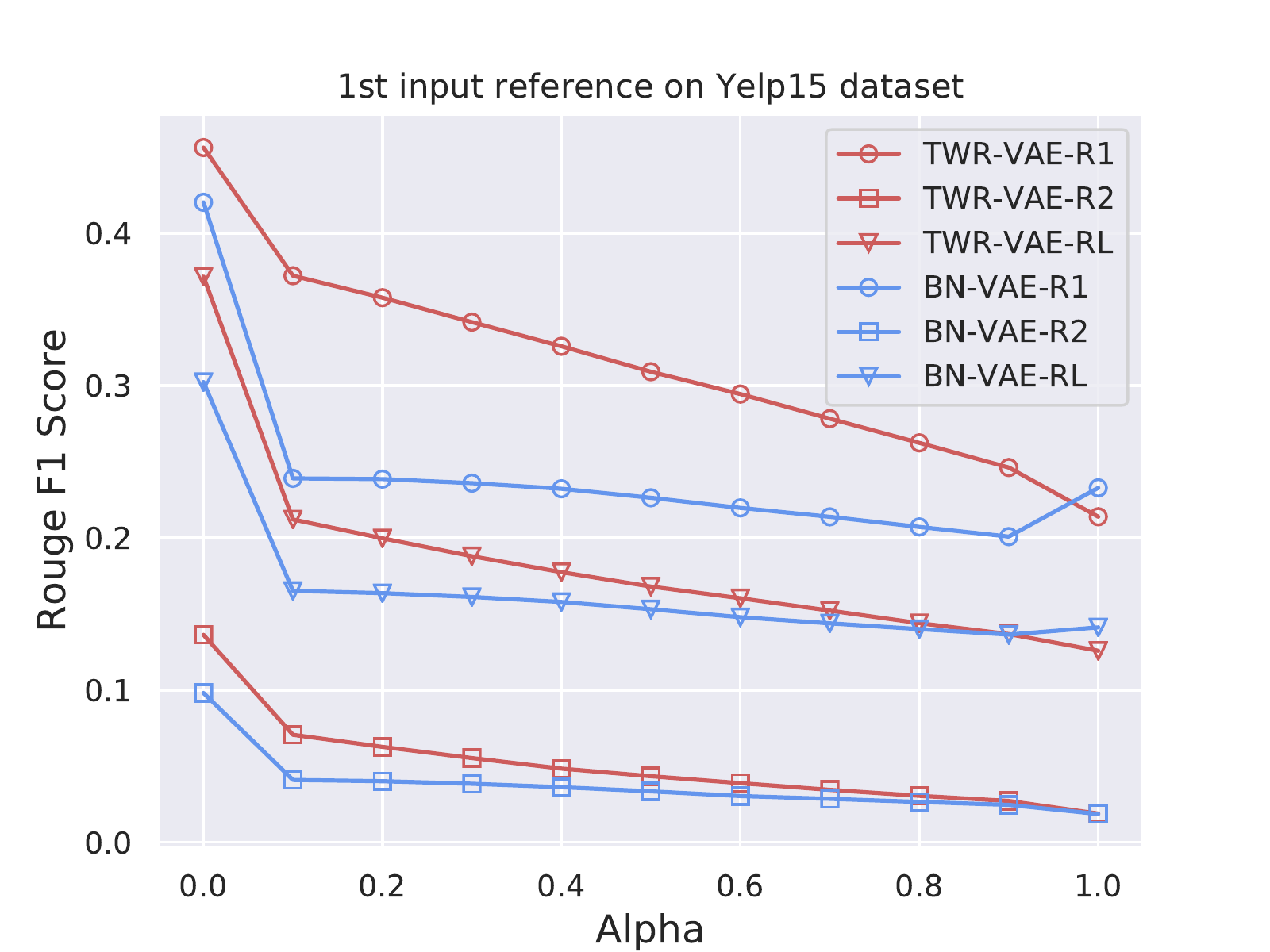}
        \caption{}
        \label{fig:R_1_yelp}
    \end{subfigure}%
    ~ 
    \begin{subfigure}[t]{0.5\textwidth}
        \centering
        \includegraphics[scale=0.5]{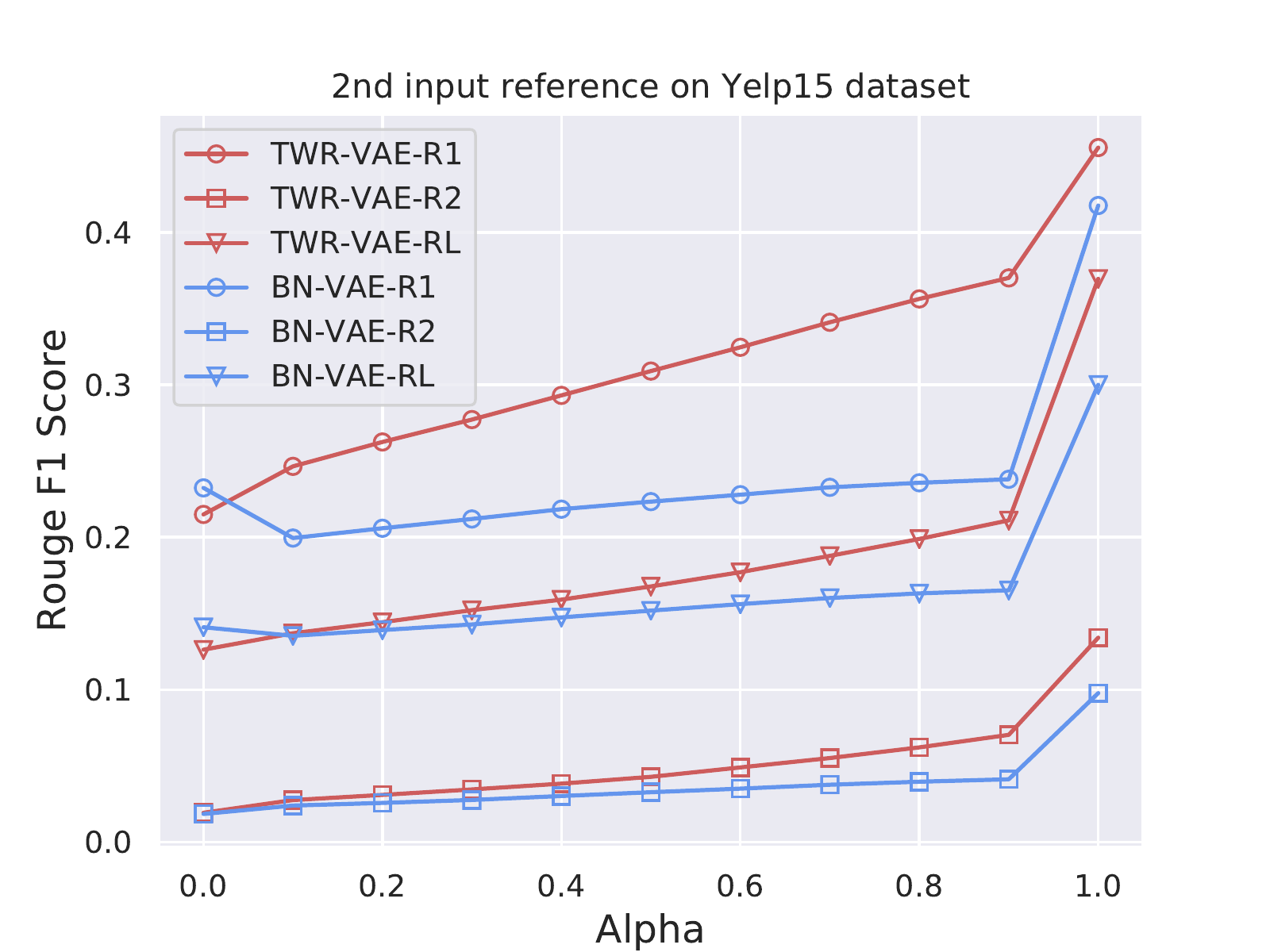}
        \caption{}
        \label{fig:R_2_yelp}
    \end{subfigure}
    \caption{The average ROUGE-1, ROUGE-2 and ROUGE-L F1 scores between two input references and 11 interpolations of each group using BN-VAE and TWR-VAE on Yelp15 test dataset (Appendix~\ref{AP:G} shows the results on Yahoo dataset).}
    \label{fig:rouge_yelp_yahoo}
\end{figure*}

\noindent\textbf{Latent representation interpolation.}~~~We perform latent representation interpolation to assess how well the latent space ($\mathbf{z}$) can be learned by TWR-VAE comparing to the strongest baseline BN-VAE. Given a pair of sentences $\mathbf{x}_1$ and $\mathbf{x}_2$, we sample their latent codes $\mathbf{z}^{\scriptscriptstyle T}_1$ and $\mathbf{z}^{\scriptscriptstyle T}_2$ from the encoder, and interpolate them with $\mathbf{z}^{\scriptscriptstyle T}_\alpha=\mathbf{z}^{\scriptscriptstyle T}_1 \cdot (1-\alpha) + \mathbf{z}^{\scriptscriptstyle T}_2 \cdot \alpha$. Table~\ref{T:sentence_yelp} shows an example outputs by varying mixture weight $\alpha$. It can be observed that our model learns representations which are more smooth than BN-VAE, where the sentences generated based on continuous samples from the latent code space preserve more consistent topical information in the neighbourhood of the path. There are less \_UNK tokens occurring in generated sentences of our model, which implies that the quality of representations learned in our model is better than ones in BN-VAE.
In addition to qualitative evaluation, we also evaluate the outputs quantitatively with ROUGE~\cite{lin-2004-rouge}, which compares the generated sentences against the human references.
Concretely, for each sentence pair, we compute the ROUGE-1, ROUGE-2 and ROUGE-L F1 scores between two input sentences (i.e., references) and each interpolation sentence. The averaged ROUGE scores over all sentence pairs in the test set versus different $\alpha$ settings are sketched in Figure~\ref{fig:rouge_yelp_yahoo}. It can observed that as the mixture weight $\alpha$ increases, the ROUGE values of our model smoothly decrease w.r.t. the first  reference and increase for the second one, showing a smooth transition of sentence interpolation. One can also note that our model has higher ROUGE scores than BN-VAE at $\alpha=0$ for reference one and at $\alpha=1$ for reference two, showing that our model is able to better learning latent representations and reconstructing the input sentences.

\subsection{Dialogue Response Generation}

In addition to language modelling, we further evaluate how well our proposed architecture could help  alleviating the problem of ``generic response'' in Dialogue Systems~\cite{huang2020challenges,wang2020dialogue}. Dialogue systems that are built upon the sequence-to-sequence (seq2seq) model were found tend to generate generic and dull responses, such as ``\textit{I don't know}'' or ``\textit{thank you}''~\cite{li2015diversity}. One effective solution is using a more flexible intermediate representation between the encoder and the decoder of a seq2seq model with the help of a VAE, which models dialogue as a one-to-many problem and, therefore, can generate less generic responses. Such VAE-based dialogue response generators, similar to~\newcite{shen2018improving}, also face the problem of posterior collapse. \newcite{zhao2017learning} first addressed this issue by proposing the conditional VAE (CVAE) model which utilises KL annealing and Bag-of-Word loss.
To test TWR-VAE on the dialogue response generation task, we extend TWR-VAE following the architecture of CVAE. 

We represent each dialogue conversation as a combination of the dialogue context $\mathbf{c}$ (context window size $J$), the response utterance $\mathbf{x}$ (the $J+1^{th}$ utterance),  and a latent representation $\mathbf{z}$ which encodes the information of the context and captures a latent distribution of valid responses. The dialogue response generation can then be defined as $P_{\boldsymbol{\theta}}(\mathbf{x}|\mathbf{c})=\int P_{\boldsymbol{\theta}}(\mathbf{x}|\mathbf{z},\mathbf{c})P_{\boldsymbol{\theta}}(\mathbf{z}|\mathbf{c})d\mathbf{z}$. Here, a vatiational posterior $Q_{\boldsymbol{\phi}}(\mathbf{z}|\mathbf{x},\mathbf{c})$ is used to approximate the true prior $P_{\boldsymbol{\theta}}(\mathbf{z}|\mathbf{c})$. The ELBO of TWR-VAE can then be written as:
\begin{align}\label{eq:our_ELBO_CVAE}
    \mathcal{L}(\boldsymbol{\theta},\boldsymbol{\phi};\mathbf{x}_i)_{\text{TWR}} &= \mathbb{E}_{Q_{\boldsymbol{\phi}}(\mathbf{z}^{J}|\mathbf{x}_i,\mathbf{c})}[\log P_{\boldsymbol{\theta}}(\mathbf{x}_i|\mathbf{z}^{J},\mathbf{c})]  - \frac{1}{J} \sum_{j=1}^{J}D_\text{KL}(Q_{\boldsymbol{\phi}}(\mathbf{z}^j|\mathbf{x}_i,\mathbf{c}) \| P_{\boldsymbol{\theta}}(\mathbf{z}^j|\mathbf{c}))\,.
\end{align}

\begin{table}[tb]
\resizebox{\columnwidth}{!}{%
  \centering
  \small
  \begin{tabular}{l|cccc|c|cccc|c}
  \toprule
  \multirow{2}{*}{Metrics} &
  \multicolumn{5}{c|}{Switchboard} &
  \multicolumn{5}{c}{Dailydialog}\\ 
     & SeqGAN & CVAE & WAE & iVAE & TWR-VAE & SeqGAN & CVAE & WAE & iVAE & TWR-VAE\\  \midrule
      BLEU-R$\uparrow$ & 0.282 & 0.295 & 0.394 & \textbf{0.427} & 0.395 & 0.270 & 0.265  &0.341 & 0.355 & \textbf{0.407}\\
      BLEU-P$\uparrow$ & \textbf{0.282} & 0.258 & 0.254 & 0.254 & 0.258 & 0.270 & 0.222 & 0.278 & 0.239 & \textbf{0.281}\\
      BLEU-F1$\uparrow$ & 0.282 & 0.275 & 0.309 & \textbf{0.319} & 0.312 &  0.270  & 0.242 & 0.306 & 0.285 & \textbf{0.333}\\
      BOW-A$\uparrow$ & 0.817 & 0.836 & 0.897 & \textbf{0.930} & 0.921 &  0.918 & 0.923 & 0.948 & 0.951 & \textbf{0.952}\\
      BOW-E$\uparrow$ & 0.515 & 0.572 & 0.627 & \textbf{0.670} & 0.654  & 0.495 & 0.543 &  0.578 & \textbf{0.609} & 0.603\\
      BOW-G$\uparrow$ & 0.748 & 0.846 & 0.887 & \textbf{0.900} & \textbf{0.900} & 0.774 & 0.811 & 0.846 & \textbf{0.872} & 0.865\\
      Intra-dist1$\uparrow$ & 0.705 & 0.803 & 0.713 & 0.828 & \textbf{0.860} & 0.747 & \textbf{0.938} & 0.830 & 0.897 & 0.921\\
      Intra-dist2$\uparrow$ & 0.521 & 0.415 & 0.651 & 0.692 & \textbf{0.849} & 0.806 & 0.973 & 0.940 & 0.975 & \textbf{0.990}\\
      Inter-dist1$\uparrow$ & 0.070 & 0.112 & 0.245 & 0.391 & \textbf{0.470} & 0.075 & 0.177 & 0.327 & \textbf{0.501} & 0.497\\
      Inter-dist2$\uparrow$ & 0.052 & 0.102 & 0.413 & 0.668 & \textbf{0.766} &  0.081 & 0.222 & 0.583 & \textbf{0.868} & 0.817\\
      
\bottomrule
  \end{tabular}
  }
  \caption{Dialogue response generation results of baselines and our model on SW and DD datasets.}
  \label{T:Dialogue}
\end{table}

\noindent\textbf{Setup.}~~We conducted experiment based on two popular benchmark datasets, namely, Switchboard (\textbf{SW})~\cite{godfrey1997switchboard}  and Dailydialog (\textbf{DD})~\cite{li2017dailydialog}. For dataset statistics, please refer to Table~\ref{T:datasets}.   
Following the implementation of CVAE, we pair each response with 10 context utterances (i.e. $J=10$) from both speakers. 
The utterance encoder is a one-layer bidirectional GRU with 300 hidden size; both the context encoder and the decoder use a one-layer GRU with 300 hidden size. The dimension of the latent variable is 200. Appendix~\ref{AP:F} shows more details.

Apart from comparing TWR-VAE to CVAE and iVAE, we further report the results of two other competitive models for dialogue response generation\footnote{\textbf{SeqGAN}:\url{https://github.com/jiweil/Neural-Dialogue-Generation}; \textbf{CVAE}:\url{https://github.com/snakeztc/NeuralDialog-CVAE}; \textbf{WAE}:\url{https://github.com/guxd/DialogWAE}; \textbf{iVAE}:\url{https://github.com/fangleai/Implicit-LVM}}, i.e.,  
\textbf{SeqGAN}~\cite{li2017adversarial} and
a conditional Wasserstein autoencoder called \textbf{WAE}~\cite{gu2019dialogwae}.
Following prior works~\cite{gu2019dialogwae,fang2019implicit}, we report performance on three evaluation metrics including: 
(1) \textit{BLEU} scores proposed by~\newcite{zhao2017learning}, which evaluates how many $n$-grams multiple generated responses match the references. \newcite{zhao2017learning} defined BLUE precision (BLEU-P) and recall (BLEU-R) as the average and maximum BLUE score, respectively, and define BLEU-F as combination of BLEU-P and BLEU-R. $n < 4$ is used in our evaluation;
(2) \textit{BOW embedding}~\cite{liu2016not}, a cosine similarity of bag-of-words embeddings between the generated response and the reference. Three different variants of BOW embedding were tested: (1) Greedy: the average cosine similarities between word embeddings of the two utterances which are greedily matched~\cite{rus2012optimal}; (2) Average: the cosine similarity between the averaged word embeddings in the two utterances~\cite{mitchell2008vector}; (3) Extreme: the cosine similarity between the largest extreme values in the word embeddings of the two utterances~\cite{pennington2014glove};
(3) \textit{Dist}~\cite{gu2019dialogwae}, which measures the diversity of the generated dialogue responses by calculating the ratio of unique $n$-grams ($n$=1,2) over all $n$-grams in the generated dialogue responses. Two types of dist (\textit{intra-dist} and \textit{inter-dist}) were tested, which are calculated within a single sampled response and between different responses, respectively. For each context in the testset, we generate 10 responses with each model and calculate aforementioned metrics averaged over all responses.

\begin{table}[tb]
\resizebox{\columnwidth}{!}{%
  \centering
  \small
  \begin{tabular}{l|l}
  \toprule
  \multicolumn{2}{l}{\textbf{Example 1}: \textbf{Topic}: Care for the elderly \textbf{Context}: to have the responsibility of putting someone in a nursing home whose mind } \\
\multicolumn{2}{l}{ was not good and could not tell you if they were being $<\text{unk}>$ or something it just would all be so different \textbf{Target}: uh - huh} \\ \hline
\multicolumn{1}{l|}{iVAE} & \multicolumn{1}{l}{TWR-VAE} \\ \hline
1. yeah uh - huh  & 1. uh - huh\\\hline
2. yeah and then go back up and go back and & 2. i see yeah and they have to go back to work and it's really sad\\ forth and go back again & \\\hline
3. right oh that makes& 3. oh gosh they don't have to worry about\\\hline
4. she's not & 4. hm how do you feel\\
\Xhline{2.5\arrayrulewidth}
  \multicolumn{2}{l}{\textbf{Example 2}: \textbf{Topic}: Relationship \textbf{Context}: what happened , john ? \textbf{Target}: nothing .} \\ \hline
\multicolumn{1}{l|}{iVAE} & \multicolumn{1}{l}{TWR-VAE} \\ \hline
1. oh , i am .  & 1. i can't sleep well .\\\hline
2. what can we do for you ? & 2. working overtime . i have been working on the weekend for a long time .\\ & i was terrified of getting a lot of headaches and i had a terrible hangover .\\\hline
3. oh what's wrong ? i didn't know anyone .& 3. oh , i am sorry . i had a terrible pain in the morning . i was so nervous . \\& i couldn ’ t find a chance to memorize the class . i was hoping to see you \\\hline
4. i have to get my phone .  & 4. well , i am not sure of it .\\

\bottomrule
  \end{tabular}
  }
  \caption{Four sample responses generated by iVAE and our model on SW (top) and DD (bottom) datasets, given context as input. Corresponding topic and target response (gold standard) are also listed. The generated utterances are different possible responses from two models. We only show the last utterance of the dialogue context here due to  space limit (the actual context window is 10).}
  \label{T:reponse_dd}
\end{table}

\noindent\textbf{Experiment Results.}~~~As shown in Table~\ref{T:Dialogue}, our model yields a stable improvement over most evaluation metrics compared to baselines. Specially, there is a significant improvement on  \textit{Dist} for SW and the \textit{BLEU} for DD, respectively, indicating that our model can generate relevant, contentful and diverse dialogue responses. There are some metrics where our model does not outperform the state-of-art baselines, but the difference is small.
We also show in Table~\ref{T:reponse_dd} two example responses generated by TWR-VAE and the best baseline iVAE. In the first example, our model can generate more topical relevant responses compared to the responses by iVAE, which implies that the latent variable of TWR-VAE can capture a hidden topic information in the dialogue conversation. In the second example, the generated responses of TWR-VAE are more diverse and contentful than the baseline, and the content of those responses can also provide more topics and 
facilitate the continuation of the conversation.

\section{Conclusion}

In this paper, in order to solve posterior collapse issue of VAE in text modelling, we propose a simple and generic model called Timestep-Wise Regularisation VAE,
which imposes the KL regularisation on the latent variables of every timestep of the encoder. Empirical results in language modelling show that our model can give better performance than all baselines while avoiding posterior collapse. Ablation studies show that the timestep-wise regularisation can be easily applied into different RNN-based VAE models and improve their performance. In addition, we evaluate the timestep-wise regularisation in dialogue response generation task, and the results suggest that our model yields better or comparable performance to the state-of-the-art and can generate relevant, contentful and diverse responses.

\section*{Acknowledgements}
This work is supported by the award made by the
UK Engineering and Physical Sciences Research
Council (Grant number: EP/P011829/1).

\bibliographystyle{coling}
\bibliography{coling2020}
\clearpage
\appendix
\section{The derivation of the ELBO (Eq. 1)}\label{AP:A}
The ELBO can be directly derivated from the marginal log likelihood $\log P_{\boldsymbol{\theta}}(\mathbf{x}_i)$:

\begin{align}
\log P_{\boldsymbol{\theta}}(\mathbf{x}_i)=& \mathbb{E}_{Q_{\boldsymbol{\phi}}(\mathbf{z} | \mathbf{x}_i)}\left[\log P_{\boldsymbol{\theta}}(\mathbf{x}_i)\right]\\
=& \mathbb{E}_{Q_{\phi}(\mathbf{z} | \mathbf{x}_i)}\left[\log \left[\frac{P_{\theta}(\mathbf{x}_i, \mathbf{z})}{P_{\theta}(\mathbf{z} | \mathbf{x}_i)}\right]\right] \\
=& \mathbb{E}_{Q_{\phi}(\mathbf{z} | \mathbf{x}_i)}\left[\log \left[\frac{P_{\theta}(\mathbf{x}_i, \mathbf{z})}{Q_{\phi}(\mathbf{z} | \mathbf{x}_i)} \frac{Q_{\phi}(\mathbf{z} | \mathbf{x}_i)}{P_{\theta}(\mathbf{z} | \mathbf{x}_i)}\right]\right] \\
=& \underbrace{\mathbb{E}_{Q_{\boldsymbol{\phi}}(\mathbf{z} | \mathbf{x}_i)}\left[\log \left[\frac{P_{\boldsymbol{\theta}}(\mathbf{x}_i, \mathbf{z})}{Q_{\boldsymbol{\phi}}(\mathbf{z} | \mathbf{x}_i)}\right]\right.}_{\underset{\text{(ELBO)}}{=\mathcal{L}(\boldsymbol{\theta}, \boldsymbol{\phi}; \mathbf{x}_i)}}+\underbrace{\mathbb{E}_{Q_{\boldsymbol{\phi}}(\mathbf{z} | \mathbf{x}_i)}\left[\log \left[\frac{Q_{\boldsymbol{\phi}}(\mathbf{z} | \mathbf{x}_i)}{P_{\boldsymbol{\theta}}(\mathbf{z} | \mathbf{x}_i)}\right]\right]}_{=D_{\text{KL}}\left(Q_{\boldsymbol{\phi}}(\mathbf{z} | \mathbf{x}_i) \| P_{\boldsymbol{\theta}}(\mathbf{z} | \mathbf{x}_i)\right)}\,,
\end{align}

\section{The reparameterisation trick for our timestep-wise latent variables}\label{AP:B}
If TWR-VAE directly samples $\mathbf{z}^t$ from the $Q_{\boldsymbol{\phi}}(\mathbf{z}^t|\mathbf{x}_i^{1:t})$, this sampling behaviour is undifferentiable. A reparameterisation trick was proposed by~\newcite{kingma2013auto} to solve this issue. Nevertheless, our TWR-VAE samples multiple $\mathbf{z}^t$ at different timesteps, and we modify the form of each $Q_{\boldsymbol{\phi}}(\mathbf{z}^{t}|\mathbf{x}_i^{1:t})$, where the mean and covariance do not directly depend on $\mathbf{z}^{t-1}$.
After using the reparameterisation trick with $\boldsymbol{\epsilon}^{t} \sim \mathcal{N}(\mathbf{0}, \mathbf{I})$, $\mathbf{z}^{t}$ can be sampled as:
\begin{align}\label{eq:sample-our-z}
    \mathbf{z}^{t}=&Q_{\boldsymbol{\phi}}(\mathbf{z}^{t}|\mathbf{x}_i^{1:t})\nonumber\\
    =&g_{\boldsymbol{\phi}}(\mathbf{h}^{t}, \boldsymbol{\epsilon}^{t}|\mathbf{x}_i^{1:t})\nonumber\\
    =&\boldsymbol{\Sigma}_{\boldsymbol{\phi}}(\mathbf{h}^{t}|\mathbf{x}_i^{1:t})^{1 / 2} \boldsymbol{\epsilon}^{t}+\boldsymbol{\mu}_{\boldsymbol{\phi}}(\mathbf{h}^{t}|\mathbf{x}_i^{1:t})\,,
\end{align}
where $\boldsymbol{\epsilon}^{t} \sim \mathcal{N}(\mathbf{0}, \mathbf{I})$, and $\mathbf{h}^{t}$ is the hidden state of the LSTM at $t$ timestep. The mean and covariance are calculated via two linear transformation layers with the $\mathbf{h}^{t}$.

\section{The derivation of the gradients optimisation of $\boldsymbol{\theta}$ and $\boldsymbol{\phi}$ (Eq. 4)}\label{AP:C}

When optimising the $\boldsymbol{\theta}$ and the $\boldsymbol{\phi}$, we use Monte Carlo method~\cite{metropolis1949monte} in order to construct a Monte Carlo estimator, which can obtain unbiased gradients of $\boldsymbol{\theta}$ and $\boldsymbol{\phi}$:
\begin{align}
&\nabla_{\boldsymbol{\theta}} \mathcal{L}(\boldsymbol{\theta},\boldsymbol{\phi};\mathbf{x}_i)\nonumber\\ &=\nabla_{\boldsymbol{\theta}}  \left(\mathbb{E}_{Q_{\boldsymbol{\phi}}(\mathbf{z}^{\scaleto{T}{3.5pt}}|\mathbf{x}_i)}\left[\log P_{\boldsymbol{\theta}}(\mathbf{x}_i|\mathbf{z}^{\scriptscriptstyle T})\right] -\frac{1}{T} \sum_{t=1}^{T}D_\text{KL}\left(Q_{\boldsymbol{\phi}}(\mathbf{z}^t|\mathbf{x}_i^{1:t})\|P(\mathbf{z}^t)\right)\right)\label{eq:expectation} \\
&=\nabla_{\boldsymbol{\theta}}  \left(\mathbb{E}_{Q_{\boldsymbol{\phi}}(\mathbf{z}^{\scaleto{T}{3.5pt}}|\mathbf{x}_i)}\left[\log P_{\boldsymbol{\theta}}(\mathbf{x}_i|\mathbf{z}^{\scriptscriptstyle T})-\frac{1}{T} \sum_{t=1}^{T}\log\frac{Q_{\boldsymbol{\phi}}(\mathbf{z}^t|\mathbf{x}_i^{1:t})}{P(\mathbf{z}^t)}\right]\right) \\
&=\mathbb{E}_{Q_{\boldsymbol{\phi}}(\mathbf{z}^{\scaleto{T}{3.5pt}}|\mathbf{x}_i)}\left[\nabla_{\boldsymbol{\theta}}\left(\log P_{\boldsymbol{\theta}}(\mathbf{x}_i|\mathbf{z}^{\scriptscriptstyle T})-\frac{1}{T} \sum_{t=1}^{T}\log\frac{Q_{\boldsymbol{\phi}}(\mathbf{z}^t|\mathbf{x}_i^{1:t})}{P(\mathbf{z}^t)}\right)\right]\\
& \simeq \frac{1}{M}\sum_{m=1}^{M} \nabla_{\boldsymbol{\theta}}\left(\log P_{\boldsymbol{\theta}}(\mathbf{x}_i|\mathbf{z}_m^{\scriptscriptstyle T})-\frac{1}{T} \sum_{t=1}^{T}\log\frac{Q_{\boldsymbol{\phi}}(\mathbf{z}_m^t|\mathbf{x}_i^{1:t})}{P(\mathbf{z}_m^t)}\right) \quad \text{where} \quad \mathbf{z}_m^{\scriptscriptstyle T} \sim Q_{\boldsymbol{\phi}}(\mathbf{z}^{\scriptscriptstyle T}|\mathbf{x}_i)\\
&=\frac{1}{M}\sum_{m=1}^{M} \nabla_{\boldsymbol{\theta}}\left(\log P_{\boldsymbol{\theta}}(\mathbf{x}_i|\mathbf{z}_m^{\scriptscriptstyle T})\right) \quad \text{where} \quad \mathbf{z}_m^{\scriptscriptstyle T} \sim Q_{\boldsymbol{\phi}}(\mathbf{z}^{\scriptscriptstyle T}|\mathbf{x}_i)\,,\label{eq:Monte_Carlo_estimator}
\end{align}
which is an unbiased Monte Carlo gradient estimator to approximate the expectation (Eq.~\ref{eq:expectation}), and
$M$ indicates the total number of times that we randomly sample $\mathbf{z}^{\scriptscriptstyle T}_m$ from the $Q_{\boldsymbol{\phi}}(\mathbf{z}_{m}^{\scriptscriptstyle T}|\mathbf{x}_i^{1:t})$ for approximation.

When applying the similar method to obtain the unbiased gradients of $\boldsymbol{\phi}$, there is an obstacle to finishing the gradients:
\begin{align}
\nabla_{\boldsymbol{\phi}} \mathcal{L}(\boldsymbol{\theta},\boldsymbol{\phi};\mathbf{x}_i)
&=\nabla_{\boldsymbol{\phi}}  \left(\mathbb{E}_{Q_{\boldsymbol{\phi}}(\mathbf{z}^{\scaleto{T}{3.5pt}}|\mathbf{x}_i)}\left[\log P_{\boldsymbol{\theta}}(\mathbf{x}_i|\mathbf{z}^{\scriptscriptstyle T})-\frac{1}{T} \sum_{t=1}^{T}\log\frac{Q_{\boldsymbol{\phi}}(\mathbf{z}^t|\mathbf{x}_i^{1:t})}{P(\mathbf{z}^t)}\right]\right) \label{eq:gradient_phi}\\
& \neq \mathbb{E}_{Q_{\boldsymbol{\phi}}(\mathbf{z}^{\scaleto{T}{3.5pt}}|\mathbf{x}_i)}\left[\nabla_{\boldsymbol{\phi}}\left(\log P_{\boldsymbol{\theta}}(\mathbf{x}_i|\mathbf{z}^{\scriptscriptstyle T})-\frac{1}{T} \sum_{t=1}^{T}\log\frac{Q_{\boldsymbol{\phi}}(\mathbf{z}^t|\mathbf{x}_i^{1:t})}{P(\mathbf{z}^t)}\right)\right]\,,
\end{align}
However, we can tackle this issue by using the reparameterisation trick proposed by~\cite{kingma2013auto}. Normally, we choose a differentiable and invertible function $g_{\boldsymbol{\phi}}(\mathbf{z},\boldsymbol{\epsilon})$ with the random variable $\boldsymbol{\epsilon}$ to replace $Q_{\boldsymbol{\phi}}(\mathbf{z}|\mathbf{x}_i)$, namely $\mathbf{z}=g_{\boldsymbol{\phi}}(\mathbf{x},\boldsymbol{\epsilon})$, where $\boldsymbol{\epsilon} \sim P(\boldsymbol{\epsilon})$ (see Eq.~\ref{eq:sample-our-z}). We choose $\mathcal{N}(\mathbf{0},\mathbf{I})$ as $P(\boldsymbol{\epsilon})$ and we can use the Monte Carlo estimator approximate Eq.~\ref{eq:gradient_phi}:
\begin{align}
    &\nabla_{\boldsymbol{\phi}}  \left(\mathbb{E}_{Q_{\boldsymbol{\phi}}(\mathbf{z}^{\scaleto{T}{3.5pt}}|\mathbf{x}_i)}\left[\log P_{\boldsymbol{\theta}}(\mathbf{x}_i|\mathbf{z}^{\scriptscriptstyle T})-\frac{1}{T} \sum_{t=1}^{T}\log\frac{Q_{\boldsymbol{\phi}}(\mathbf{z}^t|\mathbf{x}_i^{1:t})}{P(\mathbf{z}^t)}\right]\right)\nonumber\\
    &=\nabla_{\boldsymbol{\phi}}  \left(\mathbb{E}_{P(\boldsymbol{\epsilon})}\left[\log P_{\boldsymbol{\theta}}(\mathbf{x}_i|\mathbf{z}^{\scriptscriptstyle T})-\frac{1}{T} \sum_{t=1}^{T}\log\frac{Q_{\boldsymbol{\phi}}(\mathbf{z}^t|\mathbf{x}_i^{1:t})}{P(\mathbf{z}^t)}\right]\right)\\
    &=\mathbb{E}_{P(\boldsymbol{\epsilon})}\left[\nabla_{\boldsymbol{\phi}}\left(\log P_{\boldsymbol{\theta}}(\mathbf{x}_i|\mathbf{z}^{\scriptscriptstyle T})-\frac{1}{T} \sum_{t=1}^{T}\log\frac{Q_{\boldsymbol{\phi}}(\mathbf{z}^t|\mathbf{x}_i^{1:t})}{P(\mathbf{z}^t)}\right)\right]\\
    &\simeq \frac{1}{M}\sum_{m=1}^{M}\nabla_{\boldsymbol{\phi}}\left(\log P_{\boldsymbol{\theta}}(\mathbf{x}_i|\mathbf{z}_m^{\scriptscriptstyle T})-\frac{1}{T} \sum_{t=1}^{T}\log\frac{Q_{\boldsymbol{\phi}}(\mathbf{z}_m^t|\mathbf{x}_i^{1:t})}{P(\mathbf{z}_m^t)}\right)\\
    &=\frac{1}{M}\sum_{m=1}^{M}\nabla_{\boldsymbol{\phi}}\left(-\frac{1}{T} \sum_{t=1}^{T}\log\frac{Q_{\boldsymbol{\phi}}(\mathbf{z}_m^t|\mathbf{x}_i^{1:t})}{P(\mathbf{z}_m^t)}\right) \nonumber \\ & \quad \text{where} \quad \mathbf{z}_m^t=g_{\boldsymbol{\phi}}^t\left(\boldsymbol{\epsilon}_{m}, \mathbf{x}_{i}^{1:t}\right) \quad \text{and} \quad \boldsymbol{\epsilon}_m \sim \mathcal{N}(\mathbf{0},\mathbf{I})
\end{align}
Overall, the gradients of $\boldsymbol{\theta}$ and $\boldsymbol{\phi}$ of the ELBO can be re-formed as:
\begin{align}\label{eq:full_gradient_theta_phi_HR-VAE-ELBO}
    &\nabla_{\boldsymbol{\theta},\boldsymbol{\phi}}\mathcal{L}(\boldsymbol{\theta},\boldsymbol{\phi};\mathbf{x}_i)\nonumber\\
    &=\nabla_{\boldsymbol{\theta},\boldsymbol{\phi}}  \left(\mathbb{E}_{P(\boldsymbol{\epsilon})}\left[\log P_{\boldsymbol{\theta}}(\mathbf{x}_i|\mathbf{z}^{\scriptscriptstyle T})-\frac{1}{T} \sum_{t=1}^{T}\log\frac{Q_{\boldsymbol{\phi}}(\mathbf{z}^t|\mathbf{x}_i^{1:t})}{P(\mathbf{z}^t)}\right]\right)\\
    &=\mathbb{E}_{P(\boldsymbol{\epsilon})}\left[\nabla_{\boldsymbol{\theta},\boldsymbol{\phi}}\left(\log P_{\boldsymbol{\theta}}(\mathbf{x}_i|\mathbf{z}^{\scriptscriptstyle T})-\frac{1}{T} \sum_{t=1}^{T}\log\frac{Q_{\boldsymbol{\phi}}(\mathbf{z}^t|\mathbf{x}_i^{1:t})}{P(\mathbf{z}^t)}\right)\right]\\
    &\simeq \frac{1}{M}\sum_{m=1}^{M}\nabla_{\boldsymbol{\theta},\boldsymbol{\phi}}\left(\log P_{\boldsymbol{\theta}}(\mathbf{x}_i|\mathbf{z}_{m}^{\scriptscriptstyle T})-\frac{1}{T} \sum_{t=1}^{T}\log\frac{Q_{\boldsymbol{\phi}}(\mathbf{z}_{m}^t|\mathbf{x}_i^{1:t})}{P(\mathbf{z}_{m}^t)}\right)\nonumber \\ & \quad \text{where} \quad \mathbf{z}_{m}^t=g_{\boldsymbol{\phi}}^t\left(\boldsymbol{\epsilon}_{m}, \mathbf{x}_{i}^{1:t}\right) \quad \text{and} \quad \boldsymbol{\epsilon}_{m} \sim \mathcal{N}(\mathbf{0},\mathbf{I})\,,
\end{align}

\section{Training Details for Language Modelling}\label{AP:D}
We represent input data with 512-dimensional word2vec embeddings~\cite{mikolov2013distributed} and set the dimension of the hidden layers of both 1-layer encoder and decoder to 256. The dimension of the latent variable is 32. There is no gradient clipped during training. The Adam optimiser~\cite{kingma2014adam} is used for training with an initial learning rate of 1e-4 and a weight decay of 1e-5. Each sentence in a mini-batch is padded to the maximum length for that batch, and the maximum batch-size allowed is 64.

\section{The derivation of $I(\mathbf{x},\mathbf{z})$ }\label{AP:E}
\begin{align}
    &\mathbb{E}_{\mathbf{x}}[D_\text{KL}\left(Q_{\phi}(\mathbf{z}^{\scriptscriptstyle T}|\mathbf{x})\|P(\mathbf{z}^{\scriptscriptstyle T})\right)]\label{eq:old_KL}\\
    &=\mathbb{E}_{\mathbf{x}}[\mathbb{E}_{Q_{\phi}(\mathbf{z}^{\scaleto{T}{3.5pt}}|\mathbf{x})}[\log Q_{\phi}(\mathbf{z}^{\scriptscriptstyle T}|\mathbf{x})]] - \mathbb{E}_{\mathbf{x}}[\mathbb{E}_{Q_{\phi}(\mathbf{z}^{\scriptscriptstyle T}|\mathbf{x})}[\log P(\mathbf{z}^{\scriptscriptstyle T})]]\\
    &=-H(Q_{\phi}(\mathbf{z}^{\scriptscriptstyle T}|\mathbf{x}))-\mathbb{E}_{Q_{\phi}(\mathbf{z}^{\scaleto{T}{3.5pt}})}[\log P(\mathbf{z}^{\scriptscriptstyle T})]\\
    &=-H(Q_{\phi}(\mathbf{z}^{\scriptscriptstyle T}|\mathbf{x}))+H(Q_{\phi}(\mathbf{z}^{\scriptscriptstyle T}))-H(Q_{\phi}(\mathbf{z}^{\scriptscriptstyle T}))  -\mathbb{E}_{Q_{\phi}(\mathbf{z}^{\scaleto{T}{3.5pt}})}[\log P(\mathbf{z}^{\scriptscriptstyle T})]\label{eq:entropy_cross_entropy}\\
    &=I(\mathbf{x},\mathbf{z}^{\scriptscriptstyle T})+\mathbb{E}_{Q_{\phi}(\mathbf{z}^{\scaleto{T}{3.5pt}})}[\log Q_{\phi}(\mathbf{z}^{\scriptscriptstyle T})]-\mathbb{E}_{Q_{\phi}(\mathbf{z}^{\scaleto{T}{3.5pt}})}[\log P(\mathbf{z}^{\scriptscriptstyle T})]\\
    &=I(\mathbf{x},\mathbf{z}^{\scriptscriptstyle T})+D_\text{KL}(Q_{\phi}(\mathbf{z}^{\scriptscriptstyle T})\|P(\mathbf{z}^{\scriptscriptstyle T}))\label{eq:mutual_info_new_KL}\,,
\end{align}
Therefore:
\begin{align}
    I(\mathbf{x},\mathbf{z}^{\scriptscriptstyle T})&=\mathbb{E}_{\mathbf{x}}[D_\text{KL}\left(Q_{\phi}(\mathbf{z}^{\scriptscriptstyle T}|\mathbf{x})\|P(\mathbf{z}^{\scriptscriptstyle T})\right)]-D_\text{KL}(Q_{\phi}(\mathbf{z}^{\scriptscriptstyle T})\|P(\mathbf{z}^{\scriptscriptstyle T}))\,,
\end{align}

\section{Training Details for Dialogue Response Generation}\label{AP:F}
Our model follows the implementation details of the CVAE~\cite{zhao2017learning}. The size of word embedding is 200 and it is initialised from a pre-trained Glove embedding on Twitter~\cite{pennington2014glove}.  The utterance encoder is a one-layer bidirectional GRU with 300 hidden size, and both of the context encoder and the decoder use a one-layer GRU with 300 hidden size. The recognition network is 1-layer feed-forward network and prior network is 2-layer feed-forward network plus a tanh non-linearity for Gaussian prior sampling. The dimension of the latent variable is 200. The context window size $J$ is 10. The initial weights for recognition and prior networks are sampled from a uniform distribution [-0.02, 0.02]. The vocabulary size is 10,000 and all out-of-vocabulary words are defined as ``$<\text{unk}>$" token. A greedy decoding mode is used to sample dialogue responses in order to ensure that the randomness comes from the latent variables. The entire model is trained using Adam optimiser with an initial learning rate of 1e-4 and a weight decay of 1e-5. Gradient clipping is not used.

\section{Examples of the latent representation interpolation on the Yahoo test dataset}\label{AP:G}

\begin{figure*}[tb]
    \centering
    \begin{subfigure}[t]{0.5\textwidth}
        \centering
        \includegraphics[scale=0.5]{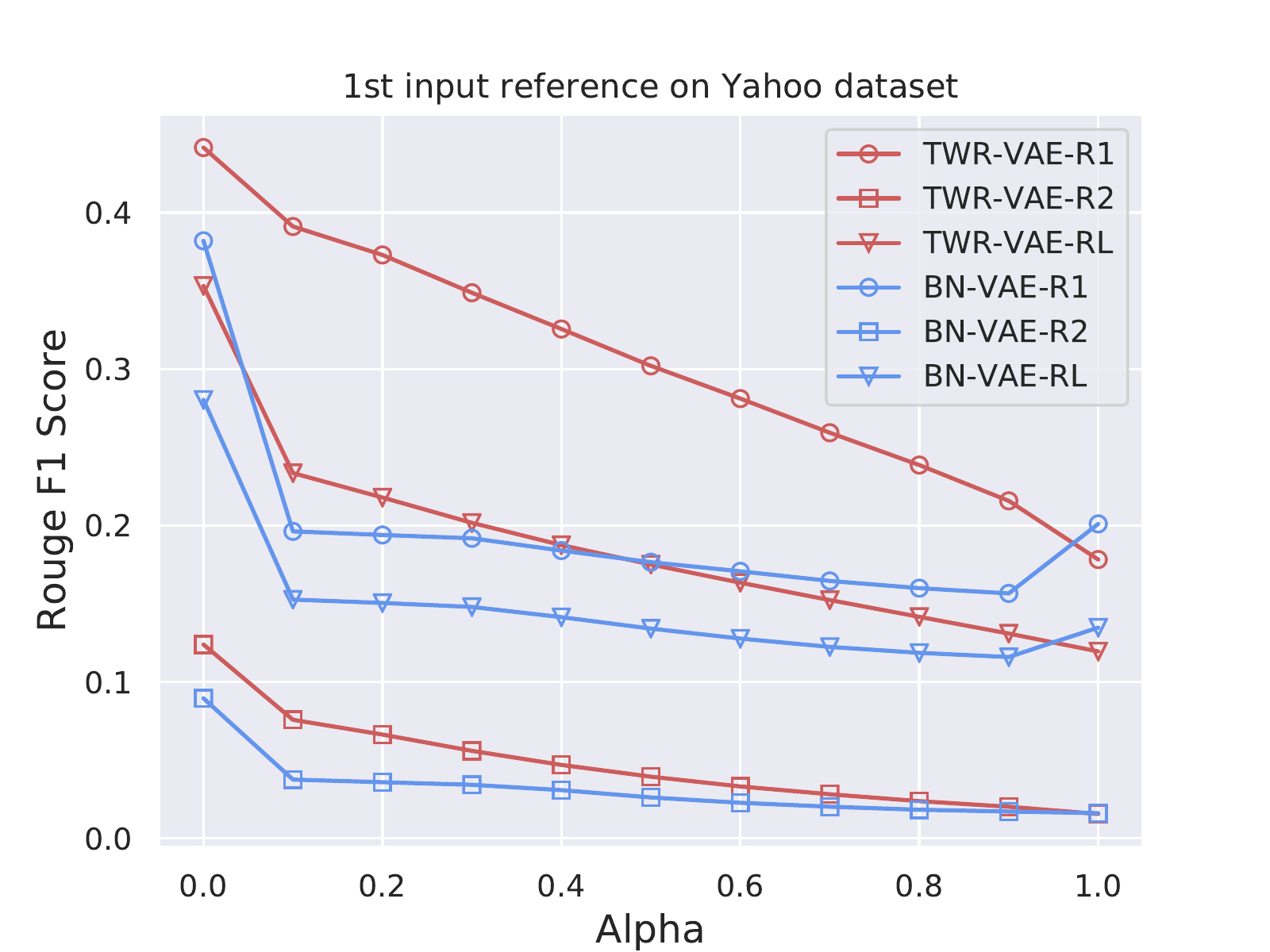}
        \caption{}
        \label{fig:R_1_yahoo}
    \end{subfigure}%
    ~ 
    \begin{subfigure}[t]{0.5\textwidth}
        \centering
        \includegraphics[scale=0.5]{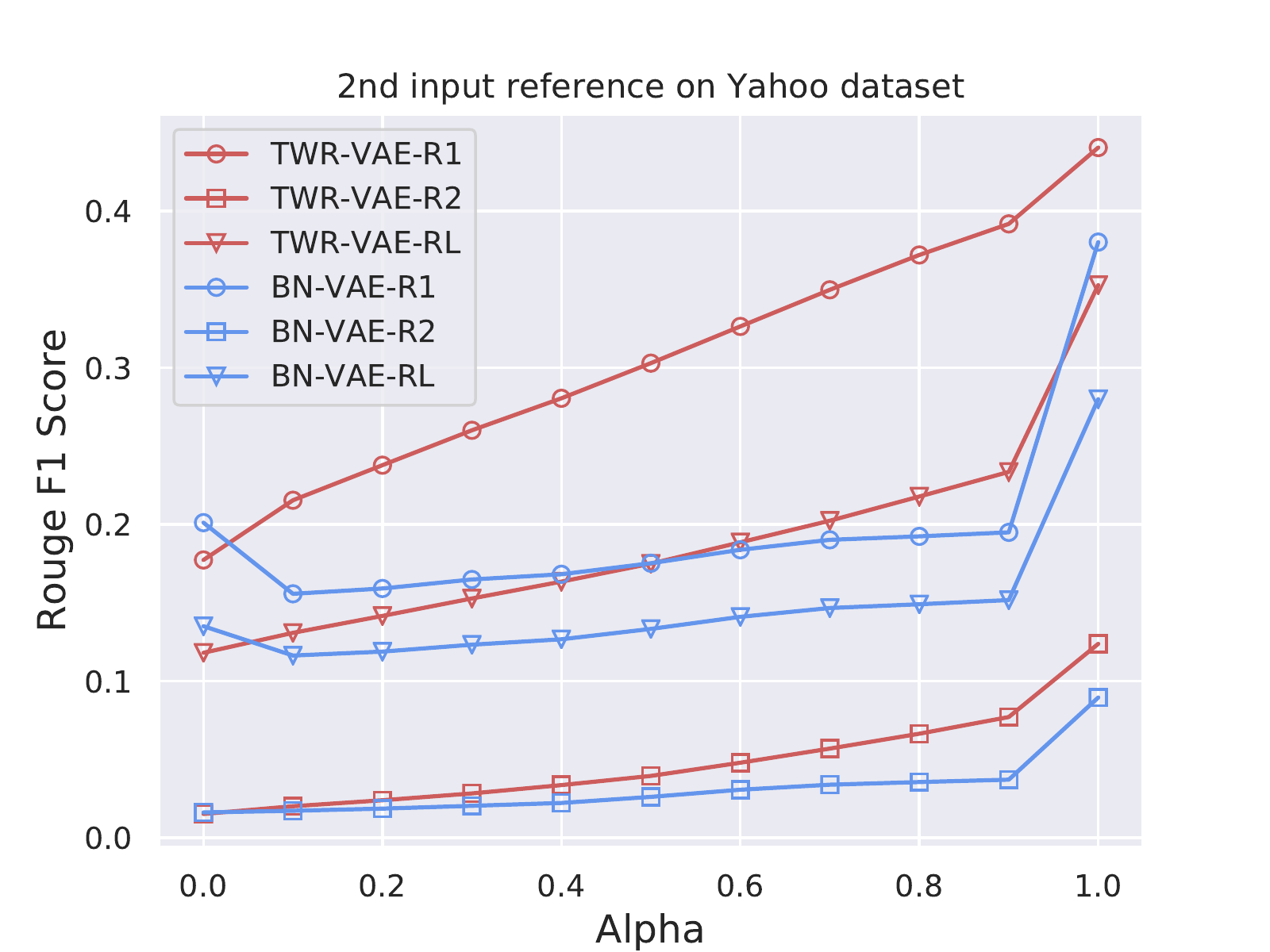}
        \caption{}
        \label{fig:R_2_yahoo}
    \end{subfigure}
    \caption{The average ROUGE-1, ROUGE-2 and ROUGE-L F1 scores between two input references and 11 interpolations of each group using BN-VAE and TWR-VAE on Yahoo test dataset.}
    \label{fig:rouge_yahoo}
\end{figure*}

There are less \_UNK tokens and repeated words occurring in the interpolated sentences generated by our model compared to BN-VAE, as shown in Table~\ref{T:sentence_yahoo}. Figure~\ref{fig:rouge_yahoo} shows that our model has higher ROUGE scores than BN-VAE at $\alpha=0$ for reference one and at $\alpha=1$ for reference two. Moreover, the ROUGE-L scores of our model are even higher than the ROUGE-1 scores of BN-VAE at $\alpha=\{0.1,0.2,0.3\}$ for reference one and at $\alpha=\{0.7,0.8,0.9\}$ for reference two. 

\begin{table*}[tb] 
\small
\resizebox{\linewidth}{!}{%
  \centering 
  \begin{tabular}{c|l|l} 
  \Xhline{2.5\arrayrulewidth}
\multirow{2}{*}{Yahoo} & Input 1 & wher can i find a poem called `` in flight '' ? it has something to do with death dunno\\
    &Input 2 &  where can i find dinosaur books for my 3 yr old son ? just check with your local library .\\ \hline
    \multirow{6}{*}{\rotatebox[origin=c]{90}{BN-VAE}} & $\alpha=0$ & can can i find a list about `` \_UNK the '' ? i is to to do with the . .\\
    &$\alpha=0.2$ & can tell me what is the name of the song on the \_UNK and the \_UNK ? i think it is a \_UNK song . \\
    &$\alpha=0.4$ & where can i find a list of all the \_UNK in the world ? i need to find a list of the \_UNK and \_UNK of the \_UNK .\\
    &$\alpha=0.6$ & where can i find a list of all the \_UNK in the world ? i need to find a list of the \_UNK and \_UNK of the \_UNK .\\
    &$\alpha=0.8$ & where can i find a list of all the \_UNK in the world ? i need to find a list of the \_UNK and \_UNK of the \_UNK .\\
    &$\alpha=1$ & where can i find a \_UNK ? free son year old son ? i go out the local library . they\\
    \Xhline{2.5\arrayrulewidth}
    \multirow{6}{*}{\rotatebox[origin=c]{90}{TWR-VAE}} & $\alpha=0$ & where can i find a pic in `` in touch attendant ? it has been to do with someone and what\\
    &$\alpha=0.2$ & in my opinion what can be done ? it 's a poem for me on myspace .com and some people have no clue \\
    &$\alpha=0.4$ & where can i find an old testament to find out how old it was ? i 'm looking at a photograph of albert einstein .\\
    &$\alpha=0.6$ & where can i find an old book for someone who has an old son ? i need to know how to do it ! !\\
    &$\alpha=0.8$ & where can i find info on my research for an anatomy book ? try these links to your local newspaper . good luck \\
    &$\alpha=1$ & where can i find info for my son year old son ? try be out your local library . good\\
    \Xhline{2.5\arrayrulewidth}
    \end{tabular}
  }
  \caption{The example of interpolating the latent representation of two input sentences using BN-VAE and TWR-VAE in Yahoo test dataset.}
  \label{T:sentence_yahoo}
\end{table*}
\end{document}